\documentclass[lettersize,journal]{IEEEtran}
\usepackage[deletedmarkup=sout,authormarkup=superscript,final]{changes} % to track changes
\usepackage{amsmath,amsfonts,amssymb}
\usepackage{amsthm}
\newtheorem{definition}{Definition}
\newtheorem{theorem}{Theorem}

\newtheorem{assumption}[definition]{Assumption}

\usepackage{algorithm}
\usepackage{algorithmicx}
\usepackage{algpseudocode}
\usepackage{array}
\usepackage[]{subfig}
\usepackage{textcomp}
\usepackage{stfloats}
\usepackage{url}
\usepackage{verbatim}
\usepackage{graphicx}
\usepackage{acro} %for acronyms
\usepackage{cite}
\usepackage{tikz, pgfplots}
\pgfplotsset{compat=1.18}
\usetikzlibrary{shapes.geometric, shapes.arrows, arrows, calc, positioning, fit, pgfplots.groupplots}
\usepgfplotslibrary{fillbetween}
\usepackage{hyperref} %for href in bib
\usepackage{enumitem}

\newcommand*{\context}[1]{x_{#1}}
\newcommand*{\arm}[1]{y_{#1}}
\newcommand*{\setcontext}[0]{\mathcal{X}}

\newcommand*{\samplet}[1]{\pi_{#1}}
\newcommand*{\meanrewardf}[2]{\mu(#1, #2)}
\newcommand*{\meanreward}[0]{\mu}

\newcommand*{\contextarmdist}[2]{\mathbb{Q} (#1,#2)}

\newcommand*{\crad}[2]{\text{conf}_{#1} (#2)}
\newcommand*{\Ind}[2]{I_{#1} (#2)}
\newcommand*{\Pind}[2]{I_{#1}^\text{pre} (#2)}
\newcommand*{\metricf}[2]{\mathcal{D} (#1, #2)}
\newcommand*{\metric}[0]{\mathcal{D}}
\newcommand*{\simspace}[0]{\mathcal{P}}
\newcommand*{\activeballs}[1]{\mathcal{A}_{#1}}
\newcommand*{\ball}[2]{B (#1,#2)}

\newcommand*{\windnot}[0]{w}

\newcommand*{\spotprice}[1]{\lambda^{S}_{#1}}
\newcommand*{\imbprice}[1]{\lambda^{I}_{#1}}
\newcommand*{\generation}[2]{g_{#1}^{#2}}
\newcommand*{\bid}[2]{f_{#1}^{#2}}
\newcommand*{\setwindbids}[0]{\mathcal{F}^\windnot}
\newcommand*{\vol}[2]{p^{#2}_{#1}}
\newcommand*{\spotpricefoc}[0]{\hat{\lambda}^S}
\newcommand*{\imbpricefoc}[0]{\hat{\lambda}^I}
\newcommand*{\generationfoc}[2]{\hat{g}_{#1}^{#2}}

\newcommand*{\lowerparams}[1]{{\boldsymbol{\theta}_{#1}}}
\newcommand*{\lowerparamsset}[0]{{\Theta}}
\newcommand*{\uncervar}[1]{z_{#1}}
\newcommand*{\uncerset}[0]{Z}

\newcommand{\algoname}[0]{Algorithm~\ref{alg:cbanditpseudodelay}} %CtxBanditBid
\acsetup{list/template=longtable, uppercase/list}
\DeclareAcronym{WPP}{
    short = WPP,
    long = wind power producer,
    tag = abbrev
}
\DeclareAcronym{TSO}{
    short = TSO,
    long = transmission system operator,
    tag = abbrev
}
\DeclareAcronym{ENTSOE}{
    short = ENTSO-E,
    long = European Network of Transmission System Operators for Electricity,
    tag = abbrev
}
\DeclareAcronym{CMAB}{
    short = CMAB,
    long = contextual multi-armed bandit,
    tag = abbrev,
    long-format=\itshape
}
\DeclareAcronym{FCR}{
    short = FCR,
    long = frequency containment reserve,
    tag = abbrev
}
\DeclareAcronym{aFRR}{
    short = aFRR,
    long = automatic frequency restoration reserve,
    tag = abbrev
}
\DeclareAcronym{mFRR}{
    short = mFRR,
    long = manual frequency restoration reserve,
    tag = abbrev
}
\DeclareAcronym{MILP}{
    short = MILP,
    long = mixed-integer linear program,
    tag = abbrev
}
\DeclareAcronym{SPDDU}{
    short = SP-DDU,
    long = stochastic program with decision-dependent uncertainty,
    tag = abbrev
}

\DeclareAcronym{windnot}{
    short = \ensuremath{w},
    long = wind power producer,
    tag = notn_part
}
\DeclareAcronym{spotprice}{
    short = \(\spotprice{}\),
    long = spot price,
    tag = notn_mdmp
}
\DeclareAcronym{imbprice}{
    short = \(\imbprice{}\),
    long = imbalance price,
    tag = notn_mdmp
}
\DeclareAcronym{generation}{
    short = \(\generation{}{}\),
    long = realized power generation at delivery time,
    tag = notn_mdmp
}
\DeclareAcronym{bid}{
    short = \(\bid{}{}\),
    long = bid price as a piecewise linear function of volume $\vol{}{}$,
    tag = notn_mdmp
}
\DeclareAcronym{vol}{
    short = \(\vol{}{}\),
    long = energy volume (MWh),
    tag = notn_mdmp
}
\DeclareAcronym{lowerparams}{
    short = \(\lowerparams{}\),
    long = lower-level program parameters,
    tag = notn_model
}
\DeclareAcronym{lowerparamsset}{
    short = \(\lowerparamsset\),
    long = set of all feasible $\lowerparams{}$,
    tag = notn_model
}
\DeclareAcronym{uncerset}{
    short = \(\uncerset\),
    long = set of all feasible ${\uncervar{}}$,
    tag = notn_model
}
\DeclareAcronym{lowermap}{
    short = \(J\),
    long = lower-level solution map,
    tag = notn_model
}
\DeclareAcronym{pardist}{
    short = $\mathbb{P} {(x,y)} $,
    long = distribution on \(\lowerparams{}\) parameterized in ${x,y}$,
    tag = notn_model
}
\DeclareAcronym{powerset}{
    short = $\mathcal{P}$,
    long = power set,
    tag = notn_model
}
\DeclareAcronym{pardistreward}{
    short = $\mathcal{H}{(x,y)}$,
    long = reward distribution parameterized in ${x,y}$,
    tag = notn_model
}
\DeclareAcronym{uncer}{
    short = $\uncervar{}$,
    long = vector of market outcome and wind power generation,
    tag = notn_model
}
\DeclareAcronym{wasser}{
    short = $\mathcal{W}$,
    long = Wasserstein distance,
    tag = notn_model
}
\DeclareAcronym{domain}{
    short = dom,
    long = Domain of a function,
    tag = notn_model
}
\begin{document}

\title{Learn to Bid as a Price-Maker Wind Power Producer}% in Short-Term Power Markets

\author{Shobhit Singhal, Marta Fochesato, Liviu Aolaritei, and Florian D\"orfler
\thanks{S. Singhal was with the Automatic Control Laboratory, ETH Z\"{u}rich, Switzerland. He is now with the Department of Wind and Energy Systems, Technical University of Denmark, Denmark. M. Fochesato and F. D\"{o}rfler are with the Automatic Control Laboratory, ETH Z\"{u}rich, Switzerland. L. Aolaritei is with the Department of Electrical Engineering and Computer Sciences, UC Berkeley, United States. (emails: \href{mailto:shosi@dtu.dk}{shosi@dtu.dk}, \href{mailto:mfochesato@ethz.ch}{mfochesato@ethz.ch}, \href{mailto:liviu.aolaritei@berkeley.edu}{liviu.aolaritei@berkeley.edu}, \href{mailto:dorfler@ethz.ch}{dorfler@ethz.ch})}% <-this % stops a space
% \thanks{footnote2}
}

\markboth{}%
{}

\IEEEpubid{}

\maketitle

\begin{abstract}
We consider the problem of a \Ac*{WPP} participating in short-term power markets, that faces significant imbalance costs due to its non-dispatchable and uncertain production. Additionally, some \acsp*{WPP} have a large enough market share to influence market prices with their bidding decisions, thereby rendering price forecasts unreliable\textemdash commonly referred to as the price-maker setting. We model this problem as a \acl*{CMAB} problem that leverages contextual information, such as market and generation forecasts, and accounts for the price-maker effect. We show that our algorithm achieves vanishing regret, compared to an omniscient oracle, ensuring convergence to optimal policy in the long run. The algorithm's performance is evaluated against various benchmark strategies using a numerical simulation of the German day-ahead and real-time markets.

% \Acp*{WPP} participating in short-term power markets face significant imbalance costs due to their non-dispatchable and variable production. While some \acsp*{WPP} have a large enough market share to influence prices with their bidding decisions, existing optimal bidding methods rarely account for this aspect. Price-maker approaches typically model bidding as a bilevel optimization problem, but these methods require complex market models, estimating other participants' actions, and are computationally demanding. To address these challenges, we propose an online learning algorithm that leverages contextual information to optimize \acs*{WPP} bids in the price-maker setting. We formulate the strategic bidding problem as a \acl*{CMAB}, ensuring provable regret minimization. The algorithm’s performance is evaluated against various benchmark strategies using a numerical simulation of the German day-ahead and real-time markets.
\end{abstract}


\section{Introduction}\label{sec:intro}
\IEEEPARstart{T}{he} world is moving towards decarbonized energy sources due to the urgent need for climate action. Wind energy forms a significant share of decarbonized energy sources, especially due to its widespread geographical availability and cost-effectiveness. Due to their technological maturity, \Acp{WPP} nowadays participate in the day-ahead market by submitting price-volume bids one day prior to the delivery. However, due to their non-dispatchable and uncertain production, \Acp{WPP} suffer from significant imbalance costs.

\begin{figure}
    \centering
    \begin{tikzpicture}
        \node[draw] (a) {Day-ahead market};
        \node[draw, below of = a, node distance=2cm] (b) {Real-time market};
        \node[above of = a] (c) {\color{black}price-volume bid};
        \draw[->] (a.south) -- node[left] {$\generation{}{\windnot}-${\color{black}${p}^{\windnot}$}} (b.north);
        \draw[->] (c.south) -- (a.north);
        \node[draw, right of = a, node distance = 3.5cm] (a1) {\color{black}${\lambda}^S {p}^{\windnot}$};
        \node[draw, right of = b, node distance = 3.5cm] (b1) {{\color{black}$\lambda^I$} $(\generation{}{\windnot}-${\color{black}${p}^{\windnot}$}$)$};
        \draw[-] (a.east) -- (a1.west);
        \draw[-] (b.east) -- (b1.west);
        \node[anchor=center, draw] (r) at ($(a1.south)!0.5!(b1.north)$) {Revenue};
        \draw[->] (a1.south) -- node[right, xshift=-0.05cm, yshift=-0.1cm] {+} (r.north);
        \draw[->] (b1.north) -- node[right, xshift=-0.05cm, yshift=0.1cm] {+} (r.south);
    \end{tikzpicture}
    \caption{A price-maker \ac{WPP} participating in the day-ahead and real-time markets. The day-ahead market clearing produces a dispatch schedule $\vol{}{\windnot}$ and the resulting imbalance $(\generation{}{\windnot}-\vol{}{\windnot})$ is settled in the real-time market, where $\generation{}{\windnot}$ denotes the realized \ac{WPP} generation. $\spotprice{}, \imbprice{}$ denote the day-ahead and real-time market prices, respectively. In the price-maker setting, the day-ahead bid affects both the dispatch volume and the clearing price. Likewise, the day-ahead dispatch affects the imbalance volume, and thus, the real-time market price.}\label{fig:market_flow}
\end{figure}
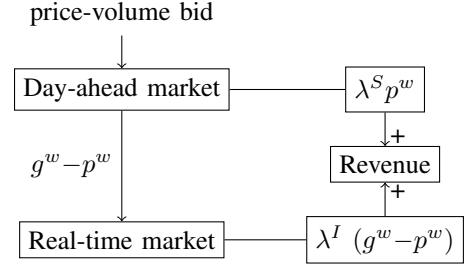

    Stochastic programming has traditionally been used to maximize \acp{WPP} revenue amidst production uncertainty~\cite{pinson_2007,morales_2010,dent_2011,zhang_2012,antonio_2013,michael_2014,li_2016,singh_2019}. These works develop optimal bidding strategies for sequential day-ahead and real-time markets under a price-taker setting, incorporating generation and market price forecasts. However, the price-taker assumption, i.e. the \ac{WPP}'s bidding decisions do not impact market prices, does not hold for all \acp{WPP}. Many European countries, such as Denmark (55\%) and Germany (22\%), have a large wind power share in their generation mix. Consequently, a large \ac{WPP} can not trust market price forecasts and needs to account for its own impact on prices, as illustrated in Fig.~\ref{fig:market_flow}. The impact on market price is especially pronounced in the intraday and real-time markets due to low trade volumes, for instance, the average proportion of the balance energy traded is $\sim1\%$, compared to the day-ahead market~\cite{netztransparenz}. However, due to uncontrollable production, the only strategic leverage available to a \ac{WPP} is due to arbitrage between market stages, akin to virtual bidding~\cite{xiao2022virtual}. For example, a \ac{WPP} expecting higher real-time price is incentivized to bid below its forecasted volume in the day-ahead market. However, excessive underbidding can raise downregulation demand, lower real-time price, and ultimately eliminate or even reverse the arbitrage benefit. Thus, for a large \ac{WPP}, an arbitrage with a small share of its production capacity can impact market prices significantly, and thus, its revenue. Clearly, in this regime, the price-taker assumption ceases to hold, and the corresponding bidding strategies are suboptimal.

To address this issue, researchers have modeled the price-maker setting as a stochastic bilevel problem, where the upper-level optimizes bids to maximize revenue, while the lower-level simulates market clearing for the chosen bid and returns clearing price and dispatch volumes. The stochasticity in market information required for simulating market clearing is handled using scenarios derived from expert knowledge or forecasting methods. The resulting optimization problem is a \ac{MILP}, and solved using off-the-shelf solvers. Notably, \cite{zugno_pool_2013} considers the setting where the \ac{WPP} is a price-taker in the day-ahead market but a price-maker in the real-time market due to relatively large imbalance volumes for a \acp{WPP}. Differently, \cite{baringo_strategic_2013} considers the setting where the \ac{WPP} is a price-maker in the day-ahead market but a price-taker in the real-time market and determines optimal bidding strategies, while ~\cite{xiao2022virtual} further considers virtual bidding. Similarly,~\cite{ding2017wfess} computes the optimal bidding strategy for a wind-storage plant, using linear decision rules for the battery in the real-time market achieving 10\% higher profit by considering price-maker effect. Further, \cite{delikaraoglou_price-maker_2015,dai2015,baringo2016risk} considers price-maker effect in both the market stages. Finally,~\cite{shafiekhah2015idrx} considers the presence of demand response, while~\cite{baringo2021virtual} optimizes strategy for a virtual power plant.\par

However, the MILP-based approaches mentioned above face several challenges. Most notably, they require extensive market information to model the lower-level market clearing, including, participants' bids and marginal costs. Unfortunately, much of this information might be private\textemdash such as marginal costs or capacities\textemdash or only revealed in future\textemdash such as participants' aggregated bids. While prior works incorporate forecasts of such information through scenario optimization, it worsens the computational complexity of the resulting \ac{MILP} due to a large number of scenario variables introduced in the lower-level market clearing problem. For example,~\cite{kraft2023stochastic} reports up to 3 hours of computation time for a single problem instance. This is not aligned with the ongoing shift of power markets towards shorter lead times, such as in intraday auctions~\cite{fortumblog}.

Online trading algorithms are a promising solution, since they continually learn and adapt optimal bidding strategies from real-time data streams. These algorithms aim to minimize the average regret, defined as the average difference between the revenue of the optimal bid in hindsight and that of the proposed bid. The absence of a need for re-training as new data becomes available, combined with inexpensive update steps, makes them computationally efficient and suitable for rapid decision-making. For example, authors in~\cite{wang2022earning,abate_learning_2024,baltaoglu2018algorithmic} employ multi-armed bandit (MAB) algorithms to optimize participation in oligopolistic markets. MAB problem assume independent repeated markets, i.e. the prior decisions do not impact future outcomes. While, this holds true for renewable producers such as \acp{WPP}, it may not apply to assets such as energy storage systems whose state of charge depends on prior bidding decisions. Reinforcement learning extends the MAB problem to account for dynamic states~\cite{ye_deep_2020}.

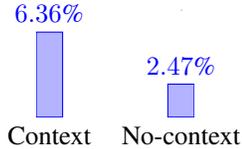
\begin{figure}
    \centering
    \begin{tikzpicture}
    \begin{axis}[scale=0.25,
        ybar,
        x axis line style = {opacity = 0},
        x=7cm,
        symbolic x coords = {Context,No-context},
        xtick=data,
        nodes near coords={\pgfmathprintnumber{\pgfplotspointmeta}\%},
        axis y line = none,
        ymin=0, ymax=8,
        tickwidth = 0pt]
        \addplot coordinates {
        (Context,6.36)
        (No-context,2.47)
        };
    \end{axis}
    \end{tikzpicture}
    \caption{Potential improvement in \ac{WPP} revenue by incorporating contextual information into the bidding strategy, compared to a context blind approach, for the proposed algorithm in Section~\ref{sec:algorithm}. The results are based on historical German market data, with the simulation details provided in Section~\ref{sec:simsetup}.}\label{fig:contextnocontext}
\end{figure}

While the above works assumed a stationary electricity market across repeated instances, each instance is impacted by exogenous variables like fuel prices, renewable production, and uncertain demand. As energy traders routinely have access to day-ahead forecasts on market status and weather conditions, we are interested in exploiting the availability of this \textit{contextual information} to make better bidding decisions, as suggested by Fig.~\ref{fig:contextnocontext}. In this direction,~\cite{ni2023contextual} employs linear contextual bandits for financial portfolio optimization, that assumes a linear relationship between observed contexts and expected outcome. While the linearity assumption simplifies the learning problem, it might be restrictive. Notably,~\cite{munoz2023online} develops a linear contextual bidding policy for a price-taker producer offering in the day-ahead and two-price settlement real-time markets. However, its applicability is limited to the specific market structure.

\noindent \textbf{Contributions. }In this work, we develop an online learning bidding algorithm that uses contextual information to compute an optimal bidding strategy for a price-maker \ac{WPP}. Specifically, the paper makes the following main contributions:
\begin{itemize}
        \item The optimal bidding problem for a price-maker is formulated as a stochastic program with a decision- and context-dependent uncertainty, agnostic to the market structure. This formulation leverages contextual information and enables the application of \ac{CMAB} algorithms.
        \item We adapt the \ac{CMAB} algorithm in~\cite{slivkins2014contextual} for the setting of short-term power markets, and show that the algorithm achieves zero regret asymptotically.
        \item We develop a simulation framework for the day-ahead and real-time markets using historical data from Nord Pool~\mbox{\cite{nordpooldata}} and ENTSO-E~\mbox{\cite{entsoe}}. To account for the price-maker effect, we propose forecasts for first order market information\textemdash such as day-ahead market revenue sensitivity\textemdash as contextual information. Finally, we evaluate our algorithm's performance against several benchmarks.
\end{itemize}
Our results show that the proposed bidding strategy yields higher cumulative revenue compared to alternative strategies, highlighting the benefits of \ac{CMAB}-based bidding strategies.

\noindent \textbf{Outline.} The rest of the paper is organized as follows. Section~\ref{sec:market} describes and models the market stages considered in this paper. Section~\ref{sec:setting} outlines the problem setting, followed by the algorithm description in Section~\ref{sec:algorithm}. Section~\ref{sec:validation} provides the numerical simulation and validation method along with results. Section~\ref{sec:conclusion} concludes the paper.

\section{Price-maker setting in short-term power markets}\label{sec:market}
Here, we model the German day-ahead and real-time markets~\cite{epexspot,regelleistung}, followed by the \ac{WPP}'s participation problem considering strategic behavior in both market stages.

\subsection{Repeated day-ahead and real-time markets}\label{sec:da_rt_market}
The day-ahead market allows market participants to buy or sell electricity for physical delivery on the following day. It consists of a batch of 24 simultaneous auctions (one for each hour of the day) held one day prior to the delivery, repeated every day. For each of the 24 hourly auctions, a participant submits a price-volume bid. Consequently, the market is ``cleared'', i.e., an optimal dispatch problem is solved that maximizes the social welfare subject to market and network constraints. Let $\bid{}{\windnot}$ be the \ac{WPP}'s day-ahead bid, for instance, a piecewise constant price-volume bid. We define $\lowerparams{}$ to include all the exogenous information affecting the day-ahead and real-time market clearings, such as day-ahead bids of other participants, balance energy provider bids, and the realized generation $\mathbf{g}$. The market operator optimizes a social welfare function $h^S(\mathbf{p}; \bid{}{\windnot},\lowerparams{})$, subject to feasiblity set $S^S(\bid{}{\windnot},\lowerparams{})$ representing network, market, and regulatory constraints\footnote{Note that in \eqref{eq:daclearing} we neglect coupling constraints, such as block bids and ramping limits, as typically done in comparable works~\cite{delikaraoglou_price-maker_2015,zugno_pool_2013}. Indeed, ramping limits and block bids are relevant in the case of thermal and hydropower plants, while they are obsolete in the presence of a large share of renewables. Thus, each hourly auction is independent of the others and the previous decisions do not impact future outcomes.}: for a single hourly auction, the corresponding optimization problem reads
\begin{subequations}\label{eq:daclearing}
    \begin{align}
        \underset{\mathbf{p}}{\max}& \quad h^S(\mathbf{p}; \bid{}{\windnot},\lowerparams{})\\
        \text{s.t.} & \quad \mathbf{1}^\top \mathbf{p} = 0 \quad ; \quad {\color{black}\spotprice{}}\label{eq:dapriceform}\\
        & \quad \mathbf{p}\in S^S(\bid{}{\windnot},\lowerparams{}),
    \end{align}
\end{subequations}

where \eqref{eq:dapriceform} enforces power balance in the day-ahead dispatch and the corresponding dual variable returns the spot price $\spotprice{}$ which is used to settle all the accepted bid volumes~\cite[Section~5.6]{boyd2004convex}. Let the entry corresponding to the \ac{WPP}'s dispatch schedule be $\vol{}{\windnot}$; then, the payment received by the \ac{WPP} is $\spotprice{}\vol{}{\windnot}$. The optimal dispatch problem in~\eqref{eq:daclearing} is a parametric program in $\bid{}{\windnot}$ and $\lowerparams{}$; thereby the corresponding primal and dual solutions are denoted as $\mathbf{p}^\star(\bid{}{\windnot},\lowerparams{})$ and $\spotprice{}(\bid{}{\windnot},\lowerparams{})$, respectively.
 
While all the participants are expected to adhere to the day-ahead schedule, \acp{WPP} deviate due to their uncertain production. The resulting imbalance, i.e., the difference between the realized dispatch $\mathbf{g}$ and the scheduled dispatch $\mathbf{p}^\star$, defines the total balance energy demand (up- or down-regulation) that is settled on the real-time market. For the supply side of the real-time market, the balance energy providers submit price-volume bids ahead of time for both up- and down-regulation. Similar to the day-ahead market, optimal dispatch problem~\eqref{eq:realtimeclearing} activates the required amount of balance energy\footnote{In the European power market, there are mainly three types of balancing energy reserves differing in their speed of response \cite{regelleistung}: \ac{FCR}, \ac{aFRR}, and \ac{mFRR}. In this work, balancing energy shall refer to the \ac{aFRR} type reserve, which contributes to the largest share of balancing energy costs. The \ac{FCR} is relatively small in comparison to the \ac{aFRR}, while the \ac{mFRR} is typically activated only in extreme cases like power plant failures.}.
\begin{subequations}\label{eq:realtimeclearing}
    \begin{align}
        \underset{\mathbf{r}}{\max}& \quad h^I(\mathbf{r}; \mathbf{p}^\star,\lowerparams{})\\
        \text{s.t.} & \quad \mathbf{1}^\top (\mathbf{g}-\mathbf{p}^\star) + \mathbf{1}^\top\mathbf{r} = 0 \quad ; \quad {\color{black}\imbprice{}}\label{eq:imbpriceform}\\
        & \quad \mathbf{r}\in S^I(\mathbf{p}^\star,\lowerparams{}),
    \end{align}
\end{subequations}

where \eqref{eq:imbpriceform} enforces real-time power balance and the corresponding dual variable represents the imbalance price $\imbprice{}$. \acp{WPP} imbalance is given by $\generation{}{\windnot}-\vol{}{\windnot}$, yielding a payment to the \ac{WPP} of ${\imbprice{}(\generation{}{\windnot} - \vol{}{\windnot})}$. $\imbprice{}(\bid{}{\windnot},\lowerparams{})$ denotes the resulting imbalance price, since the day-ahead dispatch $\mathbf{p}^\star$ is parametric in $\bid{}{\windnot},\lowerparams{}$.

\subsection{Price-maker \ac{WPP} in short-term power markets}\label{sec:bilevelform}
We are now ready to formulate the revenue optimization problem faced by a \ac{WPP} participating as a price-maker in the day-ahead and real-time markets. The total revenue from the two market stages is 
\begin{equation}\label{eq:revenue}
        \ell(z) = \spotprice{}\vol{}{\windnot} + \imbprice{}(\generation{}{\windnot} - \vol{}{\windnot}),
\end{equation}
where $z:=[\spotprice{},\vol{}{\windnot},\imbprice{},\generation{}{\windnot}]$ collects the market and generation outcomes. 
As discussed in Section~\ref{sec:da_rt_market}, market outcomes are result of the market clearings~\eqref{eq:daclearing},\eqref{eq:realtimeclearing}; thus, they depend on the bidding decision $\bid{}{\windnot}$ and exogenous variables $\lowerparams{}$. We denote the result of market clearing and generation outcome as $z^\star(\bid{}{\windnot},\lowerparams{})$.

In the price-maker setting, the \ac{WPP} maximizes the revenue \eqref{eq:revenue} by accounting for the impact of its decisions on the market outcome including clearing prices and dispatch. Mathematically, the price-maker optimal bidding problem reads as
\begin{subequations}\label{eq:toy_bilevel}
    \begin{align}
        \max\limits_{\bid{}{\windnot}\in\setwindbids} \quad & \ell(z)\label{eq:toy_toplevelobj}\\
        \text{s.t.} \quad &
        z = z^\star(\bid{}{\windnot},\lowerparams{}),\label{eq:toy_marketoutconstr}
    \end{align}
\end{subequations}
\noindent where $\setwindbids$ denotes the set of permissible bids according market regulations. Program~\eqref{eq:toy_bilevel} constitutes a bilevel problem (similar to~\cite{delikaraoglou_price-maker_2015,zugno_pool_2013,baringo_strategic_2013}), where the upper-level~\eqref{eq:toy_toplevelobj} optimizes the \ac{WPP}'s revenue and the lower-level~\eqref{eq:toy_marketoutconstr} simulates the market clearing process. Note that the bilevel structure is absent in the price-taker setting, where the market outcome remains independent of the \ac{WPP}'s bidding decision $\bid{}{\windnot}$.

\section{Problem setting}\label{sec:setting}
Consider the optimal bidding problem~\eqref{eq:toy_bilevel} for a price-maker \ac{WPP}. The exogenous variables $\lowerparams{}$ are \textit{unknown} to the \ac{WPP} at the time of bidding: information such as other participants' bids are in fact private, while variables such as wind power generation are only revealed during delivery. Conversely, contextual information, which we denote collectively as $x\in\mathcal{X}$, is typically available before bidding, for example wind power generation forecast, power consumption forecast, and fuel prices. In this paper, we seek to optimize the \ac{WPP}'s bidding decision leveraging the available contextual information.

Let $\mathbb{P}(\lowerparams{},X)$ be the joint distribution of $\lowerparams{}$ and covariate $X$. Further, let $\mathbb{P}(\lowerparams{}|X=\context{})$ be the distribution of $\lowerparams{}$ conditioned on the observed context $\context{}$. The uncertainty in $\lowerparams{}$ propagates to the \ac{WPP}'s revenue, leading to the revenue distribution\footnote{The symbol $\#$ denotes a pushforward operation. Formally, given a (measurable) map $f$ and a distribution $\mathbb P$, the pushforward of $\mathbb P$ via $f$ is defined by $(f_\# \mathbb P)(\mathcal A):= \mathbb P(f^{-1}(\mathcal A))$, for all measurable sets $\mathcal A$. In other words, if the random variable $X$ is distributed according to $\mathbb P$, then $f_\# \mathbb P$ is the distribution of the random variable $f(X)$. Finally, we note that both $\ell$ and $z^\star$ are Borel measurable, ensuring the well-posedness of the pushforward. In particular, the market clearing problems~\eqref{eq:daclearing},\eqref{eq:realtimeclearing}, as defined in~\cite{delikaraoglou_price-maker_2015}, are parametric linear programs. The corresponding primal and dual solutions are piecewise affine functions; hence, they are measurable.}:
\begin{equation}
    \contextarmdist{\bid{}{\windnot}}{\context{}}:=\ell_{\#}z^\star(\bid{}{\windnot},\cdot)_{\#}\mathbb{P}(\cdot|{\context{}}).\label{eq:revdist}
\end{equation}
In a nutshell, $\contextarmdist{\bid{}{\windnot}}{\context{}}$ represents the revenue distribution conditioned on the contextual information $\context{}$ and the \ac{WPP}'s bidding decision $\bid{}{\windnot}$.
% Comment: The bilevel vs SPDDU diagram
\begin{figure}
    \subfloat[\label{fig:model_bilevel}]{
        \begin{tikzpicture}
            \node[draw] (obj) {$\max\limits_{\bid{}{\windnot}} \ \left[ {\spotprice{} \vol{}{\windnot}} + {\imbprice{}} ({\generation{}{\windnot}} - {\vol{}{\windnot}}) \right]$};
            \node[below of = obj, node distance=2.5cm] (market) {
            \begin{minipage}{0.25\linewidth}
                \vspace{5pt}
                \begin{center}
                    Market clearing
                \end{center}
                \vspace{0pt}
            \end{minipage}};
            \draw[->] ([shift=({-0.5,-0.1})]obj.south) node[left,yshift=-0.5cm] {${\bid{}{\windnot}}$} -- ([shift=({-0.5,0})]market.north);
            \draw[->] ([shift=({0.5,0})]market.north) node[right,yshift=0.5cm] {${\spotprice{}}, {\vol{}{\windnot}}, {\imbprice{}}, {\generation{}{\windnot}}$} -- ([shift=({0.5,-0.1})]obj.south);
            \node[below of=market, node distance=1.5cm, align=center] (g) {generation $\generation{}{\windnot}$ and\\ market parameters $\lowerparams{}$};
            \draw[->] (g.north) -- ([shift=({0,0.0})]market.south);
            \node [fit=(market), draw, inner sep=-2pt] { };
        \end{tikzpicture}
    }
    \hfil
    \subfloat[\label{fig:model_sp}]{
        \begin{tikzpicture}
            \node[draw] (obj) {$\max\limits_{{\bid{}{\windnot}}\in\setwindbids}\mathbb{E} \left[\pi \right]$};
            \node[below of = obj, node distance=2.5cm, xshift=-0.1cm] (market) {
                \begin{tikzpicture}[anchor=center, font=\tiny]
                    \begin{axis}[
                    scale=0.35,
                    ymin=-0.25,
                    name=a, 
                    xticklabels={},
                    yticklabels={}]
                        \addplot[domain=-3:3]{exp(-0.5*x^2)};
                    \end{axis}
                    \node[font=\small, below of = a, yshift=1.8cm] (b) {$\contextarmdist{\bid{}{\windnot}}{\context{}}$};
                \end{tikzpicture}};
            \draw[->] ([shift=({-0.5,-0.1})]obj.south) node[left,yshift=-.2cm] {${\bid{}{\windnot}}$} -- ([shift=({-0.4,0})]market.north);
            \draw[->] ([shift=({0.6,0})]market.north) node[right,yshift=0.2cm] {$\pi$} -- ([shift=({0.5,-0.1})]obj.south);
            \node[below of=market, node distance=1.75cm, text width=1.5cm] (mi) {$x$: context};
            \draw[->] (mi.north) -- ([shift=({0,0.2cm})]market.south);
        \end{tikzpicture}
    }
    \caption{Schematic~\ref{fig:model_bilevel} refers to the bilevel formulation~\eqref{eq:toy_bilevel}, where the upper-level optimizes the \ac{WPP}'s revenue, and the lower-level represents the day-ahead and real-time markets clearing~\eqref{eq:daclearing},\eqref{eq:realtimeclearing}. The lower-level receives full information about market and wind power generation with the \ac{WPP}'s bid, and returns the market and generation outcome. Schematic~\ref{fig:model_sp} refers to the stochastic program with decision-dependent uncertainty formulation~\eqref{eq:spddu}, where the \ac{WPP} optimizes the expected revenue distributed as a parametric distribution in the \ac{WPP}'s bid and observed context.}\label{fig:model_bileveltosp}
\end{figure}
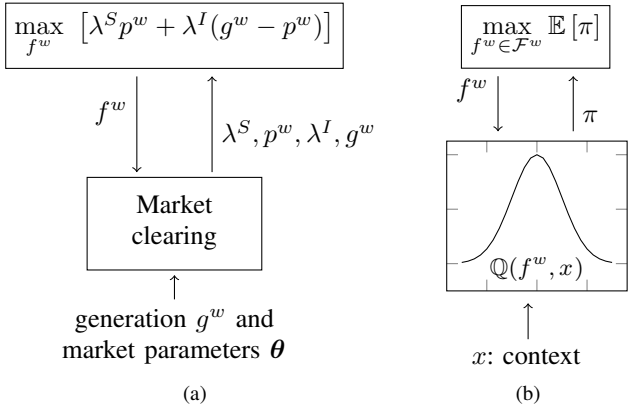
For given contextual information $x \in \mathcal{X}$, the \ac{WPP} is interested in maximizing expected revenue, i.e.,
\begin{equation}
        \bid{}{\windnot{}*}(\context{}) =\arg\max\limits_{\bid{}{\windnot}\in \setwindbids} \quad \underset{\pi\sim\contextarmdist{\bid{}{\windnot}}{\context{}}}{\mathbb{E}}[\pi].
\label{eq:spddu}
\end{equation} 

Problem \eqref{eq:spddu} constitutes a stochastic program with a (bid, context)-dependent distribution and effectively replaces the bilevel structure in \eqref{eq:toy_bilevel}. We exemplify this in Fig.~\ref{fig:model_bileveltosp}. Note that while stochastic programs with decision-dependent distributions traditionally arise in the performative prediction literature~\cite{hardt_performative_2023,drusvyatskiy_stochastic_2023}, our formulation is complicated by the additional dependence on the context. Conversely, contextual stochastic optimization~\cite{sadana_survey_2024} accounts for the effect of contexts, but considers an exogenous distribution.

The \ac{WPP} solves the bidding problem in \eqref{eq:spddu} for each bidding interval (for example, 24 hours per day). In each bidding round $t$, the following events occur in succession:
\begin{enumerate}
    \item a context $\context{t}\in \setcontext$ is revealed to the bidder,
    \item the bidder chooses a bid $\bid{t}{\windnot}\in\setwindbids$.
\end{enumerate}
Only at the end of each day, the batch of revenue ${\samplet{t}\sim \contextarmdist{\bid{t}{\windnot}}{\context{t}}}$ is revealed with expectation ${\meanrewardf{\bid{t}{\windnot}}{\context{t}}:=\mathbb{E}_{\pi\sim\contextarmdist{\bid{}{\windnot}}{\context{}}}[\pi]}$ for each hour of that day. Note that while this formulation fits the framework of (stochastic) online optimization, it differs from standard formulations due to this delayed feedback. Let $W$ be the maximum delay in receiving revenue result for any bid. For the day-ahead and real-time markets, the maximum delay is $W = 24$.

Given this setting, our goal is to derive an online learning algorithm to solve the bidding problem in \eqref{eq:spddu} under an unknown revenue distribution, while specifically accounting for the delayed feedback and leveraging contextual information. Ideally, our algorithm shall minimize the total regret $R(T)$ over the $T$ timesteps
\begin{equation}\label{eq:regret}
    R(T) := \sum_{t=0}^T \Delta_t,
\end{equation}
\noindent where $\Delta_t = \mu^\star(\context{t})-\mu({\bid{}{\windnot},\context{t}})$ is the expected instantaneous regret, and $\mu^\star(\context{t})$ is the expected revenue corresponding to some oracle bidding strategy to be determined later. Roughly speaking, minimizing the total regret over a finite-time window balances trade-off between exploration (choosing random bids to learn about revenue at a bid-context pair) and exploitation (selecting recommended bid based on the current state of knowledge). While exploration entails short-term costs, it improves the quality of subsequent exploitation. While total regret minimization is equivalent to total reward maximization, the notion of regret remains useful to analyze as it quantifies the gap relative to the oracle.\par

Note that the expected reward $\meanrewardf{\bid{}{\windnot}}{\context{}}$ can be hard to learn as it can vary arbitrarily for each bid-context pair. To guarantee that the learning problem is well-behaved for a continuous bid-context space, we rely on the following assumptions.
\begin{assumption}[Lipschitz continuity]\label{ass:continuity}
Let $\mathcal{D}$ be a distance metric in bid-context space $\simspace\subseteq\setwindbids\times\setcontext$. Then it holds that
\begin{equation}\label{eq:lipschitz}
    \lvert \meanrewardf{\bid{1}{\windnot}}{\context{1}}-\meanrewardf{\bid{2}{\windnot}}{\context{2}} \rvert \leq \mathcal{D}((\bid{1}{\windnot},\context{1}),(\bid{2}{\windnot},\context{2})).
\end{equation}
\end{assumption}
\begin{assumption}[Compactness]\label{ass:compactness}
The bid-context space $\simspace\subseteq\setwindbids\times\setcontext$ is compact.
\end{assumption}
Assumptions~\ref{ass:continuity} and~\ref{ass:compactness} are standard for online learning in continuous spaces. Intuitively, they imply that bid-context pairs that are close to each other yield similar expected rewards, and that the bid parameters and contexts lie within a finite bound. Note that no further assumptions on the problem structure are required.

\section{Online bidding algorithm}\label{sec:algorithm}
In this section, we describe the proposed bidding algorithm and present a regret analysis. Specifically, we adapt the Lipschitz contextual multi-armed bandit (LCMAB) algorithm\footnote{In the bandit literature, the term ``reward'' is standard for maximization problems; here, we use it interchangeably with ``revenue''.} in \cite{slivkins2014contextual} to delayed feedback and apply it to the bidding problem \eqref{eq:spddu}. The pseudocode is reported in Algorithm~\ref{alg:cbanditpseudodelay}.

\subsection{Algorithm description}
In this section, we first summarize the main idea of the proposed algorithm followed by a detailed description.

To solve the bidding problem~\eqref{eq:spddu}, \algoname{} iteratively explores the bid-context space focusing on regions that are statistically promising, i.e., those with high reward and frequent context arrivals. The algorithm is initialized with a bid-context space $\simspace$ defined by all the feasible bid-context pairs. At any point of time, the compact bid-context space ${\simspace\subset\mathbb{R}^n}$ (where $n$ is the sum of number of contexts and bidding decisions) is covered by balls of different radii that discretize the continuous space. At each iteration, the algorithm receives a context and estimates the upper confidence bound for each of these balls, i.e. the upper bound on expected reward of any bid-context pair inside the ball. The algorithm selects the ball with the highest upper confidence bound that contains the received context, and samples a bid from the ball. As more information is acquired, the algorithm identifies the non-promising balls and refines (i.e. creates smaller balls) the discretization of the bid-context space in the promising ones. Hence, at each iteration, the algorithm returns a bidding decision, balancing exploration and exploitation to improve its chances of selecting the optimal bid for any received context. We report the detailed pseudocode in Algorithm~\ref{alg:cbanditpseudodelay} and describe it below.

% \deleted{To solve the bidding problem~\eqref{eq:spddu}, \algoname{} estimates the expected reward associated with each bid-context pair and maintains the corresponding confidence bounds, within which the true expected reward lies with high probability. Leveraging these confidence bounds, it systematically explores the bid-context space focusing on regions that are statistically promising, i.e., those with high rewards and frequent context arrivals. Over time, the algorithm refines its discretization of the bid-context space in these regions, thereby improving its chances of selecting the optimal bid for any received context as it accumulates additional information. Algorithm~\ref{alg:cbanditpseudodelay} is described in detail below.}

A ball $\ball{c}{r}$ with center $c$ and radius $r$ in the bid-context metric space $\simspace$ with distance metric $\mathcal{D}$ (we consider L2 norm in this paper) is defined as $\ball{c}{r}~=~\{p\in~\simspace~\mid~\metricf{p}{c}\leq~r\}$. The distance metric $\mathcal{D}$ is chosen such that the diameter of $\simspace$ is 1.\par

At time $t$, the algorithm maintains an estimate $\nu_t(B)$ of the expected reward for bid-context pairs within a ball, based on the rewards $\samplet{s}$ observed in previous iterations $s\in S_t(B)$ when a bid was chosen from ball $B$. Let $n_t(B):=\lvert S_t(B) \rvert$ denote the total number of such iterations. Then,
\begin{equation}
    \nu_t(B) = \frac{1}{n_t(B)}\sum\limits_{s\in S_t(B)}\pi_s.
\end{equation}

The true expected reward for bid-context pairs in a ball $B$ lies in a confidence bound around $\nu_t(B)$. An upper confidence bound on the expected reward is referred to as pre-index
\begin{equation}
    I_t^\text{pre}(B)\overset{\Delta}{=}\nu_t(B) + r(B) + \crad{t}{B},
\label{eq:preinddef}
\end{equation}
\noindent where $\crad{t}{B}$ is the measure of uncertainty in expected reward due to finite sample approximation, and $r(B)$ denotes the radius of ball $B$ which arises due to the discretization error and Lipschitz condition~\eqref{eq:lipschitz}, defined as
\begin{equation}\label{eq:craddef}
    \crad{t}{B}\overset{\Delta}{=}\sqrt{\frac{\log T}{1+n_t(B)}}.
\end{equation}

Let $\activeballs{t}$ denote the set of all the existing balls at time $t$. An enhanced confidence bound, index $I_t(B)$ is obtained by considering pre-indices from all the balls in $\activeballs{t}$, and using the Lipschitz condition:
\begin{equation}
    I_t(B)\overset{\Delta}{=} r(B) + \min\limits_{B^\prime\in\activeballs{t}}(\Pind{t}{B^\prime}+\metric(B,B^\prime)),
\label{eq:inddef}
\end{equation}

\begin{figure}
    \centering
    \begin{tikzpicture}
        \begin{axis}[
            scale=0.6,
            xmin=0, xmax=1, ymin=0, ymax=1,
            xlabel={context}, ylabel={bid},
            xtick=\empty, 
            ytick=\empty,
            domain=0:1,
            samples=100,
            axis equal,
            clip=false
        ]
        \addplot[fill, fill opacity=0.5, thick, domain=0:360, samples=100] 
            ({0.2 + 0.3*cos(x)}, {0.55 + 0.3*sin(x)});
        \node at (axis cs: 0.25,0.9) {\textbf{A}};
        
        \addplot[thick, domain=0:360, samples=100] 
            ({0.75 + 0.2*cos(x)}, {0.75 + 0.2*sin(x)});
        \node at (axis cs: 0.6,0.95) {\textbf{D}};
        
        \addplot[fill, fill opacity=0.1, thick, domain=0:360, samples=100] 
            ({0.85 + 0.2*cos(x)}, {0.5 + 0.2*sin(x)});
        \node at (axis cs: 1,0.3) {\textbf{C}};

        \addplot[fill, fill opacity=0.3, thick, domain=0:360, samples=100] 
            ({0.65 + 0.2*cos(x)}, {0.25 + 0.2*sin(x)});
        \node at (axis cs: 0.4,0.25) {\textbf{B}};

        \addplot[fill, fill opacity=0.2, thick, domain=0:360, samples=100] 
            ({0.65 + 0.1*cos(x)}, {0.75 + 0.1*sin(x)});
        \node at (axis cs: 0.75,0.85) {\textbf{E}};

        \addplot[dashed, thick] coordinates {(0.9, 0) (0.9, 1)};
        \node at (axis cs: 0.9,-0.07) {$\context{t}$};

        \addplot[red, fill, fill opacity = 0.5, thick, domain=0:360, samples=100]
            ({0.9 + 0.1*cos(x)}, {0.8 + 0.1*sin(x)});
        \node at (axis cs: 1,0.92) {\textbf{F}};
        \addplot[red, only marks, mark=*, mark size=2] coordinates {(0.9,0.8)};
        \end{axis}
    
        \begin{axis}[
            scale=0.6 ,
            hide x axis, hide y axis,
            xmin=0, xmax=1, ymin=-0.05, ymax=1.2,
            domain=0:1,
            samples=100,
            at={(0,3.6cm)},
            width=8.5cm,
            height=2.75cm]
            \addplot[thick, blue] {exp( -(x-0.75)^2/0.03 )};
        \end{axis}
    \end{tikzpicture}
    \caption{Illustration of Algorithm~\ref{alg:cbanditpseudodelay} in a two-dimensional bid-context space. Circles represent balls, with lighter shades indicating more observed samples and thus closer to satisfying activation rule. When context $\context{t}$ arrives, balls C and D are relevant. If C has a higher index value than D, a bid (red point) is sampled from D on the dashed line. Since D meets the activation condition, a new ball F is activated. The blue curve shows the context arrival distribution, guiding finer discretization in dense regions.}\label{fig:cmabillustration}
\end{figure}
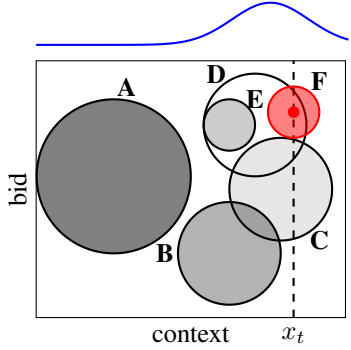

where $\mathcal{D}(B,B^\prime)$ denotes the distance between the ball centers. The algorithm's procedure is divided into two phases: predict and update. Let the current set of balls be as shown in Fig.~\ref{fig:cmabillustration}, which is used as an illustration of the algorithm's procedure in a two-dimensional bid-context space. In the prediction phase, it first receives a context $\context{t}$ (Line~\ref{alg:line5}). Then, it finds relevant balls (Line~\ref{alg:line8}) that contain the received context in their domain (balls C and D in Fig.~\ref{fig:cmabillustration}). A region of the bid-context space $\simspace{}$ can be covered by two balls of different radii with the smaller ball taking priority due to finer discretization. Thus, the domain of a ball is the remaining subset after excluding overlaps with smaller balls:
\begin{equation}
    \text{dom}(B,\activeballs{t})\overset{\Delta}{=}B\backslash\left(\cup_{B^\prime\in\activeballs{t}:r(B^\prime)<r(B)}B^\prime\right).
\end{equation}

The algorithm chooses the ball with the highest index value (optimism in the face of uncertainty) among the relevant balls (ball D in Fig.~\ref{fig:cmabillustration}, Line~\ref{alg:line7}). This is also called the selection rule in Algorithm~\ref{alg:cbanditpseudodelay}. A random bid from the selected ball is returned (Line~\ref{alg:line8}). The algorithm receives a batch of rewards at the end of the prediction phase.\par

During the update phase, the algorithm incorporates the newly observed batch of rewards and updates the index values (Line~\ref{alg:line10}). It tests whether the uncertainty due to finite sample approximation is less than the discretization error of the ball (equal to its radius due to Assumption~\ref{ass:continuity}, Line~\ref{alg:line12}). If this activation condition is met, the algorithm creates smaller balls in this region (ball F in Fig.~\ref{fig:cmabillustration}). This is called the activation rule in Algorithm~\ref{alg:cbanditpseudodelay} (Line \ref{alg:line12}-\ref{alg:line15}).\par

\begin{algorithm}[t]
    \caption{Contextual bandits for delayed feedback}
    \begin{algorithmic}[1]
        \State \textbf{Input}: Bid-context space $\simspace$
        \State \textbf{Initialize}: $B_0\leftarrow B(p,1), p\in\simspace$
        \State \hspace{40pt} $\mathcal{A}\leftarrow \{B_0\};\ n_0(B_0)=0;\ \nu_0(B_0)=0$
        \Procedure{Main Loop: for each batch $b$}{}
            \State $t^\prime=(b-1)W+1$
            \For{$t=t^\prime\ldots t^\prime+W-1$} (\textbf{Predict phase})
                \State Input context $x_t$\label{alg:line5}
                \State relevant balls $\leftarrow \{B\in\mathcal{A}_{t^\prime}: x_t \in \text{dom}(B,\mathcal{A}_{t^\prime})$\} \label{alg:line6}
                \State $B_t\leftarrow\arg\max_{B\in\text{relevant}}\Ind{t^\prime}{B}$ (\textbf{Selection rule})\label{alg:line7}
                \State $f^w_t\leftarrow$ any bid such that $(f^w,x_t)\in\text{dom}(B,\mathcal{A}_{t^\prime})$\label{alg:line8}
            \EndFor
            \State Observe batch payoff $\pi_i,\ \forall i=t^\prime \ldots t^\prime+W-1$\label{alg:line10}
            \For{$t=t^\prime\ldots t^\prime+W-1$} (\textbf{Update phase})~\label{alg:line11}
                \If{$\crad{t}{B_t}\leq r(B_t)$ and \\
                \hspace{4em}$(f^w_t,x_t)\in \text{dom}(B_t,\mathcal{A}_t)$ \label{alg:line12}} (\textbf{Activation rule})
                    \State $B^\prime\leftarrow B((f^w_t,x_t), \tfrac{1}{2}r(B_t))$
                    \State $\mathcal{A}_t\leftarrow\mathcal{A}_{t-1}\cup\{B^\prime\}; n(B^\prime)=\text{reward}(B^\prime)=0$\label{alg:line15}
                \EndIf
                \State $n(B_t)\leftarrow n(B_t)+1$; $\text{rew}(B_t)\leftarrow\text{rew}(B_t)+\pi_t$
            \EndFor
        \EndProcedure
    \end{algorithmic}
    \label{alg:cbanditpseudodelay}
\end{algorithm}

\subsection{Regret analysis}\label{sec:regretanalysis}
        As mentioned in Section~\ref{sec:setting}, our aim is to minimize the total regret with respect to a chosen oracle producing an expected revenue $\mu^\star(\context{})$. Specifically, we define oracle as the bidding strategy that maximizes the expected revenue for a given context \( \context{} \), assuming knowledge of the expected revenue associated with each bid-context pair. The corresponding expected revenue reads as
\begin{equation}
    \mu^\star(\context{}) =\max\limits_{\bid{}{\windnot}\in\setwindbids}\meanrewardf{\bid{}{\windnot}}{\context{}}.
\end{equation}

Intuitively, the oracle represents the optimal bidding decision based on the same observed contexts available to the decision-maker, including potentially noisy forecasts. An alternative oracle definition could consider an enhanced context observation, such as perfect forecasts, enabling an analysis of the impact of forecast quality on the total regret~\cite{kirschner2019stochastic}. For the chosen oracle, the following theorem provides an upper bound on the total regret incurred by \algoname{}.
\begin{theorem}[Regret bound] 
    Consider the \ac{CMAB} problem with stochastic payoffs and delayed feedback. \algoname{} achieves vanishing average regret
    \begin{equation*}
        \tfrac{R(T)}{T}\leq \mathcal{O}\left(T^{\tfrac{-1}{d_c+2}}\log T+WT^{\tfrac{-3}{d_c+2}}\right),
    \end{equation*}
    where $W$ is the maximum delay (or batch size), and $d_c$ is the r-zooming dimension.\label{theorem:regret}
\end{theorem}
% This paragraph translates the above result in simple language, and suggests a takeaway for the rest of the paper.
The proof which is an extension of the proof in~\cite{slivkins2014contextual} is reported in Appendix~\ref{app:proofregret}. The r-zooming dimension $d_c$, which is defined in Appendix~\ref{app:proofregret}, can be thought of as the effective dimension of the space of near-optimal bids corresponding to a specific context, which is at most equal to the dimension of the bid-context space $\simspace$.

Theorem~\ref{theorem:regret} suggests that the average regret decreases with increase in time horizon $T$, thereby approaching zero asymptotically. This means that the algorithm will learn to make optimal decisions almost surely with time. Moreover, the average regret increases with the maximum delay $W$, as the algorithm is unable to benefit from the reward feedback of its latest actions.

\section{Numerical validation}\label{sec:validation}
In this section, we construct a bid-context space to employ the algorithm for the optimal bidding problem. Further, we develop a market simulator for the day-ahead and real-time markets to validate the proposed algorithm against benchmark strategies. The data used for numerical validation is provided by Nordpool~\cite{nordpooldata} and ENTSO-E Transparency Platform~\cite{entsoe}.

\subsection{Simulation setup}\label{sec:simsetup}
% comment: describe the WPP
The considered price-maker \ac{WPP} is a fictitious trader that manages trade for wind turbines in the area serviced by the \ac{TSO} 50Hertz in Germany which accounts for about 20GW of installed capacity out of the 68GW total installed wind power capacity in Germany (January 2024). We describe the simulation details below.

\smallskip

\textbf{Day-ahead and real-time markets simulation}: We simulate the day-ahead auction clearing for a strategic bid of the \ac{WPP} using the historical aggregated bidding curves. We assume that the \ac{WPP} had bid competitively in the past, i.e., using the forecast bidding strategy defined in Section~\ref{sec:benchstrat}. We identify the corresponding bid, replace it with the alternative strategy bid, leading to transformed aggregated bidding curves. The spot price and dispatch volume are simply found at the intersection of these curves, neglecting any changes due to linked products such as block bids.

The imbalance price is simulated for a modified system imbalance volume using a black box approach based on the historical imbalance price and system imbalance volume data. The imbalance price depends on multiple factors, including balance energy bids and the system imbalance volume. The bid prices, in turn, are influenced by factors like fuel price, daily average spot price, and generation mix in the TSO area, which we assume fixed for a day. We then estimate a daily linear relationship between system imbalance volume and imbalance price, giving us an estimate of $\eta^I_j$, the imbalance price sensitivity to system imbalance volume for day $j$. Then, for a change of $\Delta$ in the system imbalance, the modified imbalance price is obtained by $\lambda^I+\eta^I \Delta$.

\smallskip

\textbf{Set of bidding strategies}: For the chosen market stages of the day-ahead and real-time markets, the sole decision variable is the day-ahead bid. In the German day-ahead market, bids are submitted as piecewise linear price-volume functions. For simplicity, we restrict the bid price to the marginal cost of wind power, which is considered to be zero. The remaining decision is the bid volume, which is permitted to deviate by at most $\Delta \vol{}{\windnot}$ from a reference strategy\footnote{A reference strategy allows the practitioner to leverage present knowledge and avoid unnecessarily poor decisions.}, chosen in this work to be the forecast generation volume. This results in a set of price-volume functions, expressed as
\begin{equation}
    \setwindbids:=\left\{f: [0,p]\rightarrow 0 \mid p\in[\generationfoc{}{\windnot}-\Delta\vol{}{\windnot},\generationfoc{}{\windnot}+\Delta\vol{}{\windnot}]\right\}.
\end{equation}

\smallskip

\textbf{Contextual information}: Apart from the usual power generation and market price forecasts, first order information representing price influence is important for a price-maker producer. We assume the availability of the following forecasts:

\begin{enumerate}[label=(\alph*)]
    \item Wind power generation forecast ($\generationfoc{}{\windnot}$)
    \item Spot price forecast ($\spotpricefoc$)
    \item Spot price sensitivity to bid volume ($\hat{\eta}^S$)
    \item Imbalance price forecast ($\imbpricefoc$)
    \item Imbalance price sensitivity to system imbalance ($\hat{\eta}^I$)
\end{enumerate}

The wind power generation forecast is readily accessible from ENTSO-E~\cite{entsoe}, however, the rest of the forecasts are emulated by adding noise to the estimates obtained from historical data. For instance, $\imbpricefoc = \imbprice{} + t*\sigma + \xi$, \, $t\sim T_\nu$, where $T_\nu$ denotes the Student's t-distribution with degree of freedom $\nu$, and $\xi,\sigma$ denote the location and scale parameters, respectively. The choice of the t-distribution is inspired by its widespread use in financial trading literature due to its heavier tails that allows to model the impact of outliers~\cite{rachev2007financial}.\par

Next, we engineer a single feature using the wind power generation, spot price, and spot price sensitivity forecasts. Consider the sensitivity of the day-ahead market revenue to bid volume:
\begin{equation}
    \gamma = \frac{\mathrm{d}(\spotprice{}\vol{}{\windnot})}{\mathrm{d}\vol{}{\windnot}} = \spotprice{} + \vol{}{\windnot}\eta^S,
\end{equation}
where $\eta^S=\frac{\mathrm{d}\spotprice{}}{\mathrm{d}\vol{}{\windnot}}$ represents the sensitivity of the spot price to bid volume. The corresponding forecast is obtained by substituting the respective quantities with the corresponding forecasts, i.e.:
\begin{equation}
    \hat{\gamma} = \spotpricefoc + \generationfoc{}{\windnot}\hat{\eta}^S.
\end{equation}

The resulting set of three forecast quantities ($\hat{\gamma}, \imbpricefoc, \hat{\eta}^I$) and one bidding decision (bid volume $\vol{}{}$) defines the compact bid-context space $\simspace\subset\mathbb{R}^4$, required as input to Algorithm~\ref{alg:cbanditpseudodelay}. Compactness of $\simspace$ is ensured by bounding the bidding decision through a finite deviation limit $\Delta\vol{}{\windnot}$ and estimating empirical bounds on contexts using historical data\footnote{The resulting bid-context space is the hypercube \(\simspace := [0, 1]^4\) after normalizing data to the interval [0, 1].}. 
% These bounds are designed to cover the majority of typical cases rather than all possible values, as a smaller bid-context space reduces exploration, thereby lowers regret. The objective is to develop a bidding strategy that offers effective decision support in most practical scenarios.

The algorithm's regret decreases with lower bid-context dimensionality (Theorem~\ref{theorem:regret}). Thus, while including more contexts improves the oracle strategy, it slows convergence and increases regret. Hence, selecting only relevant context variables is crucial. Domain knowledge can aid in engineering informative features that reduce dimensionality while retaining essential information.

\smallskip
\textbf{Reward}: To facilitate interpretation, reward is defined as the revenue difference between the proposed bidding algorithm and a reference bidding strategy. For instance, negative reward implies underperformance compared to the reference strategy. We choose as reference the forecast bidding strategy, as defined in Section~\ref{sec:benchstrat}. From a practitioner's perspective\textemdash where revenue from a reference strategy is not observable\textemdash the reward can be defined simply as the realized revenue, since the proposed algorithm is designed to maximize reward.

\subsection{Benchmark strategies}\label{sec:benchstrat}
In this section, we define popular bidding strategies that are later used as benchmarks for performance of the proposed algorithm.

\smallskip

\textbf{Oracle}: It refers to the bidding strategy corresponding to the oracle defined in Section~\ref{sec:regretanalysis}. Let $O:\setcontext\rightarrow\setwindbids$, then
\begin{equation}\label{eq:oracle}
    O(\context{}) = \arg\max\limits_{\bid{}{\windnot}\in\setwindbids}\meanrewardf{\bid{}{\windnot}}{\context{}},
\end{equation}
Since $\meanreward$ is unknown, we compute an estimate using the finite amount of historical data available. Detailed procedure is mentioned in Appendix~\ref{app:oracle}.
\smallskip

\textbf{Forecast bidding}: It refers to the competitive bid, i.e., forecast production volume at marginal price and is a common benchmark strategy~\cite{cao2020drl,antonio_2013}.

\smallskip

\textbf{D-1 prediction}: As outlined in Section~\ref{sec:bilevelform}, previous works model the optimal bidding problem as a bilevel program, where the lower-level represents market clearing. This formulation requires complete market information which is not available ex-ante. A natural forecasting approach is to use the corresponding market information from the previous day's market clearing. D-1 prediction is often used in power markets due to high temporal correlation~\cite{9899762,guo_forecast_2021,mitridati_bayesian_2018,wolff2017short}. We adopt the resulting bidding strategy as another benchmark, where the bid volume is given by
\begin{equation}
    \bid{t}{\windnot} = \arg\max\limits_{\bid{}{\windnot}\in\setwindbids} \ell(z^\star(\bid{}{\windnot},\lowerparams{t-24})),
\end{equation}
where $\lowerparams{t-24}$ denotes the previous day's market information.

\textbf{Linear decision rule}: Linear decision rule as suggested in~\cite{munoz2023online,munoz2020feature} is a popular approach for contextual decision making. Specifically, the bid volume is represented by a linear function of the observed contexts, $p=\generationfoc{}{\windnot} + q^\top \context{}+b$, where $q,b$ denotes weights and bias, respectively. The linear decision rule is trained on a rolling window of training set denoted by $\tilde{\mathcal{T}}(t)$ containing the latest $\lvert \tilde{\mathcal{T}(t)}\rvert$ auction results. We present numerical results for an optimized rolling window length of 150 days (3600 hourly auctions). The following optimization program returns the optimal weights $q_t$ for bidding round $t$.
\begin{subequations}
    \begin{align}
        \max_{q_t} \ & \sum_{i\in\tilde{\mathcal{T}}(t)} (\spotprice{i} - \imbprice{i})\context{i}^\top q_t\\
    \text{s.t.} \ & -\Delta\vol{}{\windnot} \leq \context{i}^\top q_t \leq \Delta\vol{}{\windnot}\quad \forall i\in\tilde{\mathcal{T}}\label{eq:lincons},
    \end{align}
\end{subequations}
where the objective function is obtained by substituting the linear decision rule in~\eqref{eq:revenue} and~\eqref{eq:lincons} enforces the maximum allowed deviation from forecast volume.

\subsection{Results}\label{sec:results}
This section presents the numerical results. We simulate the performance of all the bidding strategies from July 2022 to March 2024, resulting in a horizon length $T=15252$ auctions, corresponding to 24 hourly auctions per day. The results are obtained using Python 3.10 on a personal computer with an 8-core Intel i7-1165G7 processor and 16 GB RAM. The computation time for \algoname{} is on average 0.1 seconds per bid, which is negligible given that a trader needs 24 bids per day. Moreover, the experiments are conducted for parameters mentioned in Table~\ref{tab:simpar}, unless specified. \par

\begin{table}
    \caption{Default simulation parameters with context noise parameters $\sigma$ and $\xi$ defined with respect to normalized data.}\label{tab:simpar}
    \centering
    \begin{tabular}{c|c}
        $\Delta\vol{}{\windnot}$ & 250MW (1.25\%)\\
        $\sigma$ & 0.05 (5\%)\\
        $\xi$ & 0.0 (0\%)\\
        $\nu$ & 5\\
        $W$ & 24
    \end{tabular}
\end{table}

\begin{figure}[t]
    \centering
    \includegraphics[width=0.9\linewidth]{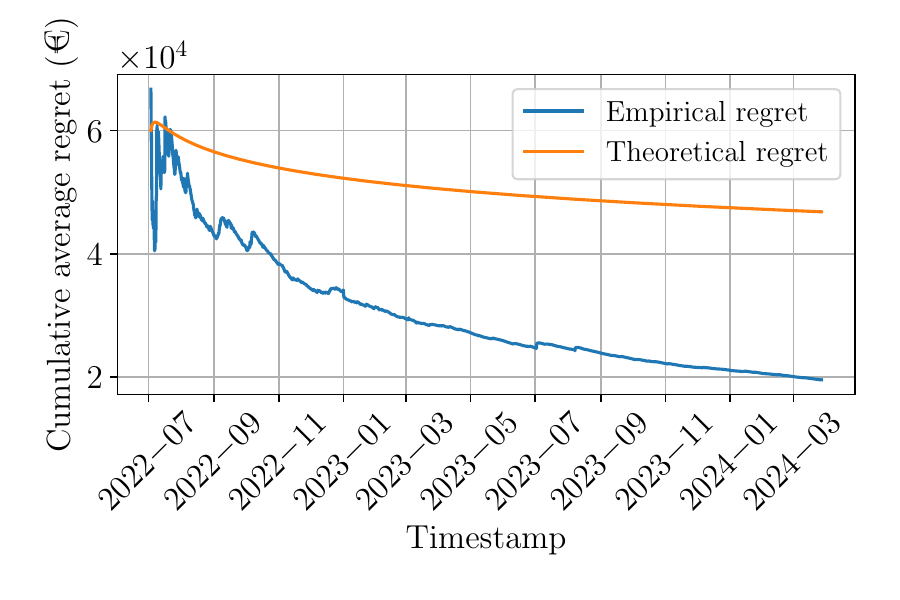}
    \caption{Evolution of the empirical and theoretical cumulative average regret with time.}\label{fig:numexp_1_1_avgimp}
\end{figure}
Fig.~\ref{fig:numexp_1_1_avgimp} shows the evolution of the empirical and theoretical cumulative average regret $R(t)/t$ with time for the proposed bidding algorithm. The theoretical regret refers to the upper bound in Theorem~\ref{theorem:regret}, which is verified by the numerical observations. In the initial iterations, the empirical regret exceeds the theoretical upper bound since the bound holds only in expectation.\par

\begin{figure}[t]
    \centering
    \includegraphics[width=0.9\linewidth]{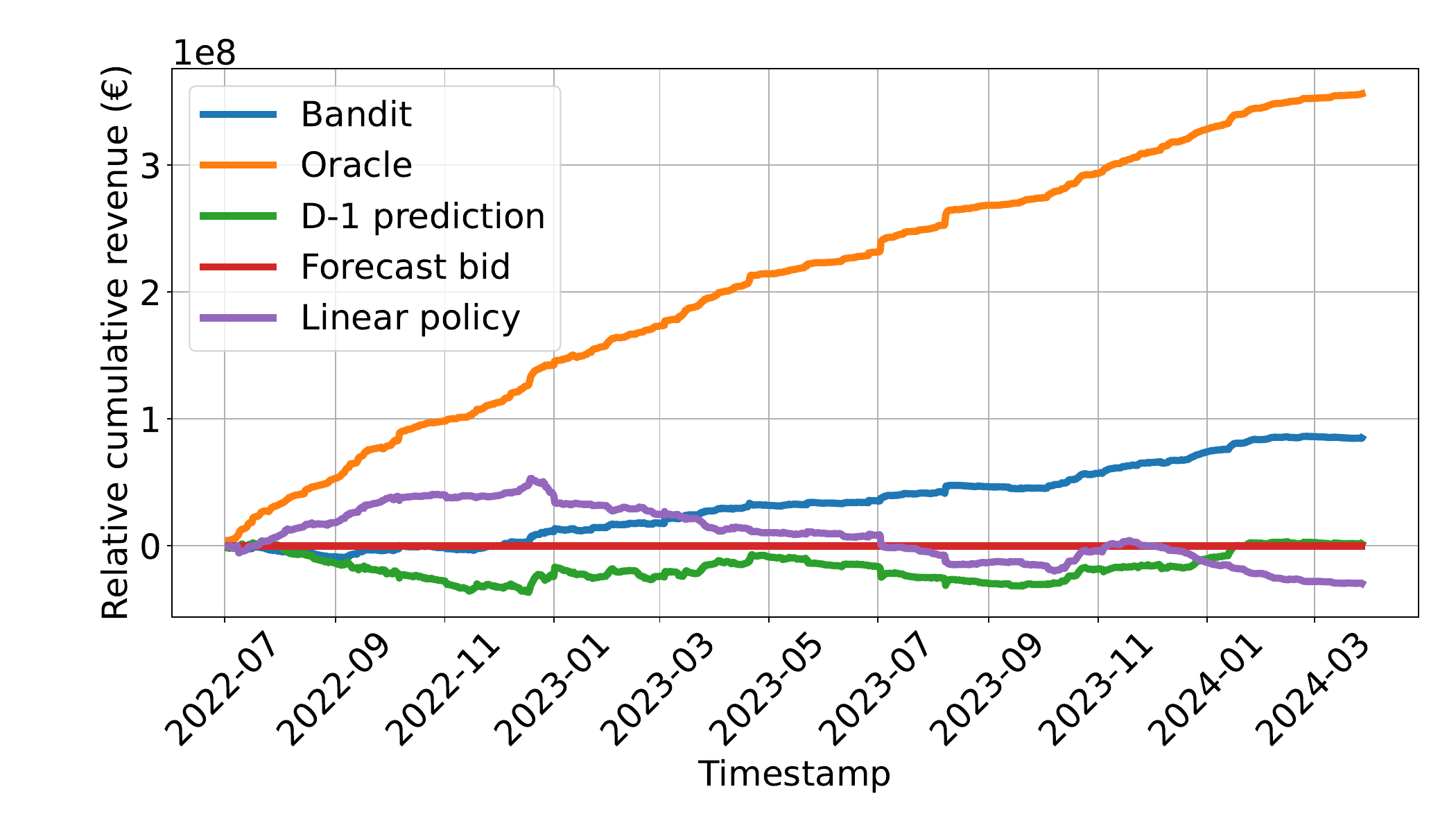}
    \caption{Cumulative revenue corresponding to different bidding strategies relative to that of the forecast bidding strategy.}\label{fig:bench_cumrev}
\end{figure}
Fig.~\ref{fig:bench_cumrev} shows the cumulative revenue for all the considered strategies, relative to the results for the forecast bidding strategy (in red). Oracle represents the theoretical upper bound on the performance of any contextual strategy (Bandit and Linear policy). Bandit underperforms initially due to exploration and achieves better performance as it accumulates data to outperform the other benchmark strategies. \added{In contrast, although the Linear policy initially exhibits strong performance due to the availability of richer contextual information, its performance diminishes over time. This decline can be attributed to the exceptionally high and volatile imbalance prices observed in 2022 ---  driven by gas market stress and transitional effects following the pricing revision implemented in June 2022. When combined with the assumption of fixed-variance forecast noise, these conditions resulted in more accurate imbalance price predictions compared to those in 2023 and 2024.} Further, the D-1 prediction often underperforms, possibly due to over reliance on preceding day's market data.

\begin{figure}[t]
    \centering
    \begin{tikzpicture}
        \begin{axis}[
            xbar,
            scale only axis,
            legend image code/.code={\draw [#1] (-0.1cm,-0.05cm) rectangle (0.15cm,0.1cm);},
            scale=1,
            width=0.12\textwidth,
            height=0.3\textwidth,
            axis y line = none,
            symbolic y coords = {\mbox{Oracle ($5.8\%$)},\mbox{Bandit ($1.4\%$)}, \mbox{D-1 prediction ($0.0\%$)}, \mbox{Forecast bid ($0.0\%$)}, \mbox{Linear policy ($-0.5\%$)}},
            nodes near coords,
            axis x line = none,
            tickwidth = 0pt,
            bar width=8pt,
            legend style={at={(0,1.15)},xshift=-1cm,anchor=north west,legend columns=2, draw=none,fill=none}
        ]
            \addplot coordinates {
                (410000,\mbox{Oracle ($5.8\%$)}) (407400,\mbox{Bandit ($1.4\%$)}) (403800,\mbox{D-1 prediction ($0.0\%$)}) (405500,\mbox{Forecast bid ($0.0\%$)}) (401846,{\mbox{Linear policy ($-0.5\%$)}})
            };
            \addlegendentry{Day-ahead}
        \end{axis}

        \begin{axis}[
            xbar,
            scale only axis,
            legend image code/.code={\draw [#1] (-0.1cm,-0.05cm) rectangle (0.15cm,0.1cm);},
            width=0.12\textwidth,
            height=0.3\textwidth,
            yshift=0.3cm,
            xshift=-0.7cm,
            y tick label style={xshift=-12mm},
            scale=1,
            axis x line=none,
            y axis line style = {opacity = 0},
            nodes near coords,
            symbolic y coords = {\mbox{Oracle ($5.8\%$)}, \mbox{Bandit ($1.4\%$)}, \mbox{D-1 prediction ($0.0\%$)}, \mbox{Forecast bid ($0.0\%$)}, \mbox{Linear policy ($-0.5\%$)}},
            tickwidth = 0pt,
            bar width=8pt,
            legend style={at={(-1,1.15)},xshift=-0.1cm,yshift=-0.3cm,anchor=north west,legend columns=2,draw=none,fill=none}
        ]
            \addplot[red, fill=red!50] coordinates {
                (13611,\mbox{Oracle ($5.8\%$)}) (-1729,\mbox{Bandit ($1.4\%$)}) (-3648,\mbox{D-1 prediction ($0.0\%$)}) (-5433,\mbox{Forecast bid ($0.0\%$)}) (-3678,\mbox{Linear policy ($-0.5\%$)})
            };
            \addlegendentry{Real-time}
        \end{axis}

    \end{tikzpicture}
    \caption{Average revenue from the day-ahead and real-time markets for all the considered strategies. The relative improvement in average revenue from both markets is mentioned in front of each bidding strategy. Both markets have a separate x-axis for better visibility.}
    \label{fig:dabal_revenue}
\end{figure}
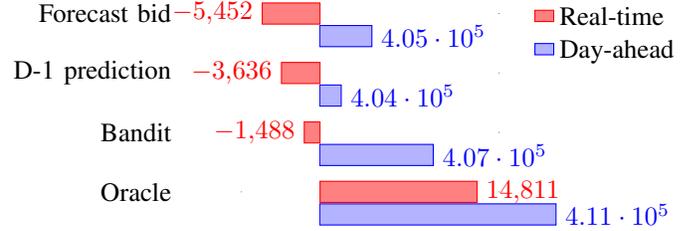

Fig.~\ref{fig:dabal_revenue} shows the split of the average revenue between the day-ahead and real-time markets and the combined percentage improvement. In the German single-price real-time market, the pricing mechanism incentivizes participants with imbalance opposite to the system imbalance~\cite{bunn2021statistical}. The positive real-time market revenue for oracle indicates that the optimal bidding strategy can capitalize on this incentive. Compared to benchmark strategies, Bandit strategy achieves higher revenue across both market stages ($1.4\%$ combined), demonstrating its ability to perform arbitrage while accounting for the price influence\textemdash particularly in the real-time market. The D-1 prediction and Linear policy fail to perform effective arbitrage, where performance in the day-ahead market is compromised for the real-time market.\par

\begin{figure}[t]
    \centering
    \begin{tikzpicture}
        \begin{groupplot}[
            grid=both,
            group style={group size=2 by 1, horizontal sep=0pt},
            scale=0.5,
            ylabel={Average revenue (\texteuro{})},
            xlabel style={yshift=0mm, align=center}
        ]
        \nextgroupplot[
            xlabel={$\Delta\vol{}{\windnot}$ (MWh)},
            xtick align=center,
            ymax=4.38e5, ymin=3.4e5,
            xtick={250,500,750},
            ytick align=center,
        ]
        \addplot[sharp plot, mark=*] coordinates {(250,405200) (500,396800) (750,372100)};
        \addplot[sharp plot, mark=triangle, mark options={solid, fill=blue}] coordinates {(250,401500) (500,381800) (750,344100)};
        \addplot[orange, sharp plot, mark=*] coordinates {(250,425500) (500,431400) (750,433400)};

        \nextgroupplot[
            xlabel={Context space\\dimensions},
            ylabel={},
            yticklabels={},
            ymax=4.38e5, ymin=3.4e5,
            xtick align=center,
            ymajorticks=false,
            legend style={at={(0.5,0)}, anchor=south, font=\scriptsize},
        ]
        \addplot[orange, sharp plot, mark=*] coordinates {(1,412419) (2,423372) (3,425554)};
        \addlegendentry{Oracle}
        \addplot[black, sharp plot, mark=*] coordinates {(1,407359) (2,410627) (3,406026)};
        \addlegendentry{Bandit ($T$)}
        \addplot[black, sharp plot, mark=triangle] coordinates {(1,402453) (2,400045) (3,394501)};
        \addlegendentry{Bandit $\left(T/2\right)$}

        \end{groupplot}
    \end{tikzpicture}
    \caption{Impact of maximum bid volume deviation from forecast $\Delta\vol{}{\windnot}$ (size of bid space) and context space dimensions (size of context space) on the average revenue achieved by \algoname{} for half $(T/2)$ and full simulation length $(T)$.}\label{fig:contextarmeffect}
\end{figure}
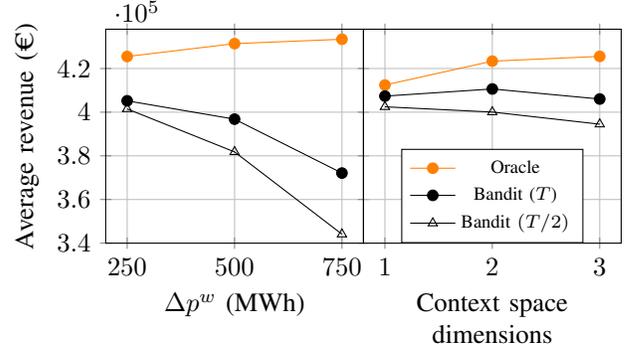

\begin{figure}
    \centering
    \begin{tikzpicture}
        \begin{axis}[
            grid=both,
            group style={group size=1 by 1, horizontal sep=0pt},
            scale=0.45,
            ylabel={Average revenue (\texteuro{})},
            xlabel style={yshift=0mm},
            xlabel={Maximum delay $W$},
            xtick align=center,
            ytick align=center,
            legend style={at={(0.5,0.5)}, anchor=center, font=\scriptsize}
            ]
            \addplot[sharp plot, mark=*] coordinates {(1,406970) (6,406565) (12,405885) (24,405672)};
            \addlegendentry{Bandit}
            \addplot[sharp plot, mark=*, violet] coordinates {(1,398802) (6,398676) (12,398452) (24,398168)};
            \addlegendentry{Linear policy}
        \end{axis}
    \end{tikzpicture}
    \caption{Impact of maximum delay on average revenue for contextual bidding strategies.}\label{fig:batcheffect}
\end{figure}

Fig.~\ref{fig:contextarmeffect} shows the impact of the maximum bid volume deviation from forecast ($\Delta \vol{}{\windnot}$) and the context space dimensions on Bandit strategy's performance. Greater freedom to deviate increases the scope of poor decisions, thereby reduces revenue-particularly in the early phase $(T/2)$ when the algorithm has not yet sufficiently explored the bid-context space $\simspace$. However, the improvement in oracle revenue suggests that, in the long run, revenue for Bandit strategy is expected to improve. Similar trend is seen for the context space dimensions for similar reasons. With both of these figures, we showcase the trade-off between the long-term and short-term performance present in bandit algorithms. 

Further, Fig.~\ref{fig:batcheffect} shows the decrease in performance with increase in maximum delay for Bandit strategy (which is consistent with Theorem~\ref{theorem:regret}) and Linear policy, however, the impact is not very significant. The D-1 prediction and Forecast bid strategies are not affected by feedback delay as they do not rely on market results for their bidding decisions.

\begin{figure}[t]
    \centering
    \begin{tikzpicture}
        \begin{groupplot}[
            grid=both,
            group style={group size=2 by 1, horizontal sep=0pt},
            scale=0.45,
            ylabel={Average revenue (\texteuro{})},
            xlabel style={yshift=0mm}
        ]
        \nextgroupplot[
            xlabel={$\xi$ (\%)},
            xtick align=center,
            ytick align=center,
            ymin=395000,
            ymax=428000
        ]
        % \addplot[sharp plot, mark=*] coordinates {(1,407382) (6,407262) (12,407182) (24,405382)};
        \addplot[sharp plot, orange, mark=*] coordinates {(0,423469) (10,423469) (20,423469) (40,423469) };
        \addplot[sharp plot, mark=*] coordinates {(0,405316) (10,405208) (20,405318) (40,405719) };
        % \addplot[sharp plot, red, mark=*] coordinates {(0,402907) (5,402907) (10,402907) (15,402907) };
        \addplot[sharp plot, red] coordinates {(0,400110) (10,400110) (20,400110) (40,400110) };
        \addplot[sharp plot, purple, mark=*] coordinates {(0,397698) (10,397548) (20,397578) (40,397324) };

        \nextgroupplot[
            axis y line*=right,
            xlabel={$\sigma$ (\%)},
            ylabel={Average revenue (\texteuro{})},
            xtick align=center,
            legend style={at={(0,1)}, anchor=south, font=\scriptsize},
            ymin=395000,
            ymax=428000
        ]
        \addplot[sharp plot, orange, mark=*] coordinates {(0,425421) (5,423469) (10,420361) (20,416145) (40,413358) };
        \addlegendentry{Oracle}
        \addplot[sharp plot, mark=*] coordinates {(0,405958) (5,405316) (10,404185) (20,402431) (40,401109)};
        \addlegendentry{Bandit}
        \addplot[sharp plot, red] coordinates {(0,400110) (5,400110) (10,400110) (20,400110) (40,400110) };
        \addlegendentry{Forecast bid}
        % \addplot[sharp plot, purple, mark=*] coordinates {(0,400215) (5,400215) (10,400215) (20,400215) (40,400215)) };
        % \addlegendentry{D-1 prediction}
        \addplot[sharp plot, violet, mark=*] coordinates {(0,397918) (5,398168) (10,398425) (20,398801) (40,399281) };
        \addlegendentry{Linear policy}
        \end{groupplot}
    \end{tikzpicture}
    \caption{Impact of context bias and noise on average revenue.}\label{fig:noiseffect}
\end{figure}

In Fig.~\ref{fig:noiseffect}, we investigate the impact of context bias and noise on the average revenue across contextual bidding strategies. Forecast bid and D-1 prediction strategies do not use context, and thus are not impacted by context noise or bias. \added{Moreover, neither of the contextual bidding strategies is affected by context bias, as the Bandit accounts for it through normalization (see Appendix~\ref{app:oracle}), while the Linear policy captures it via the intercept term.} However, increased context noise reduces Bandit strategy's performance. Interestingly, the Linear policy demonstrates greater robustness to noise and approaches the performance of the Forecast bid as context noise increases. This occurs because the policy's weights are reduced to satisfy the maximum deviation constraint~\eqref{eq:lincons}, effectively leading to predictions close to the forecast. Though, this also reflects an inherent limitation of linear models in estimating effective decisions under feasiblity constraints.

\section{Conclusion}\label{sec:conclusion}
In this paper, we proposed an online learning bidding algorithm for a price-maker \ac{WPP}, that leverages contextual information. A key contribution of this work is the alternative formulation of the optimal bidding problem for a price-maker producer, where the revenue distribution depends on both bidding decisions and contextual information, enabling application of \ac{CMAB} algorithms. The approach was validated through a simulation built using real market data, demonstrating the effectiveness of our approach over alternative strategies.\par

This work highlights several directions for future research. In this study, the reward distribution is assumed to be fixed; however, markets can change significantly over time. Therefore, investigating methods~\cite{Xu_Dong_Li_He_Li_2020,Zeng_Wang_Mokhtari_Li_2016,Suk_Kpotufe_2023,li2021detecting} to adapt to distributional shifts would be a valuable contribution. Further, the proposed algorithm imposes minimal assumptions on the structure of the parametric reward distribution; however, incorporating reasonable structural assumptions could significantly improve learning rates~\cite{krause2011contextual,majzoubi2020efficient,chu2011contextual}. Additionally, while we assume that other participants are competitive and act as price-takers, this may not always hold in practice. Thus, a valuable extension would be to consider an oligopolistic market~\cite{Sessa_Bogunovic_Krause_Kamgarpour,Zamir_2009}. Finally, expanding the market stages by including intraday markets is a natural extension of the work.

\section*{Acknowledgments}

Liviu Aolaritei acknowledges support from the Swiss National Science Foundation through the Postdoc.Mobility Fellowship (grant agreement P500PT\_222215).

\bibliographystyle{IEEEtran}
\bibliography{bibliography}

{
\appendices

\section{Oracle implementation}\label{app:oracle}
We discuss the practical implementation of Oracle~\eqref{eq:oracle} in this section. Due to a finite amount of data available in practice, it is not feasible to compute the best response for every context $\context{}\in\mathcal{X}$. Thereby, we discretize the context space and compute Oracle strategy for each discretized context.

For the numerical results presented in the paper, we have $\mathcal{X}:= [0, 1]^3$ after data normalization. Let the discretized context set be $\hat{\mathcal{X}}:=[0, 0.1, \dots, 1]^3$, and the set of observed context vectors be $\tilde{\mathcal{X}}:= \{\context{t}\}_t$. Let the discretized bid set be $\hat{\setwindbids}:=[0, 0.1, \dots, 1]$, where 0 and 1 represent deviations of $-\Delta\vol{}{\windnot}$ and $\Delta\vol{}{\windnot}$, respectively, from generation forecast.

We estimate the average revenue for each bid-context pair using a brute force methodology, and find the best bid for every discrete context vector. To obtain the data samples corresponding to a discretized context vector $\hat{x}$, we project the set of observed contexts onto the set of discretized contexts. The projection mapping is denoted by $\Pi_{\hat{\mathcal{X}}}(\context{t}):=\left\{\hat{x}\ |\ \hat{x}\in \arg\min_{\context{}\in\hat{\mathcal{X}}} \lVert \context{} - \context{t} \rVert^2 \right\}$. Then, let $\mathcal{D}_{\hat{x}}:= \left\{ t\ |\ \hat{x}\in\Pi_{\hat{\mathcal{X}}}(\context{t}),\, \context{t}\in\tilde{\mathcal{X}} \right\}$ denote the set of data samples corresponding to the discretized context $\hat{\context{}}$. Algorithm~\ref{alg:oracle} describes the methodology.
\begin{algorithm}
    \caption{Oracle strategy estimation}\label{alg:oracle}
    \begin{algorithmic}
        \State Input: Dataset $\tilde{\lowerparamsset}:=\{\lowerparams{t}\}_t,\ \tilde{\mathcal{X}}:=\{\context{t}\}_t$
        \For{$\hat{\context{}} \in \hat{\mathcal{X}}$}
            \State $max \leftarrow -\infty$
        \For{$\bid{}{\windnot} \in \hat{\setwindbids}$}
            \State $m \leftarrow \sum_{t\in\mathcal{D}_{\hat{x}}} l(z^\star(\bid{}{\windnot}, \lowerparams{t}))$
            \If{$m>max$}
                \State $max \leftarrow m$
                \State $\mathcal{O}(\hat{x}) \leftarrow \bid{}{\windnot}$
            \EndIf
        \EndFor
        \EndFor
    \end{algorithmic}
\end{algorithm}

\section{Proof of Theorem~\ref{theorem:regret}}\label{app:proofregret}
The following proof is an extension of ~\cite{slivkins2014contextual} to the delayed feedback setting. The notation used in this analysis is defined in Section~\ref{sec:algorithm}. Let reward $\pi\sim \mathbb{Q}$, with expectation $\mu$ and support $[a,b]$. For our analysis, we assume \(\lvert a - b \rvert \leq 0.5\); however, the analysis remains general for any support length, with the confidence radius increasing with the support length.\par

Then, using Hoeffding's inequality and following a procedure similar to Claim 4 in~\cite{slivkins2014contextual}, we get
\begin{equation}\label{eq:ineq_2}
    P\left(\lvert \nu_t(B) - \mu(B)\rvert \leq r(B)+\crad{t}{B}\right)\geq 1-T^{-2}.
\end{equation}
This means that the absolute deviation of the finite sample approximation of a ball $B$'s reward from its true expectation is upper bounded by $r(B)+\crad{t}{B}$ with high probability. When inequality~\eqref{eq:ineq_2} holds for the complete run of the algorithm, it is referred to as a clean run. From here on, we will assume clean run and present a deterministic analysis. Thus, the presented regret bound holds in high probability.

Recall the expected regret $\Delta(\arm{},\context{})\overset{\Delta}{=}\mu^\star(\context{})-\mu(\arm{},\context{})$ for a bid-context pair $(\arm{},\context{})$, where $\mu^\star(x)=\max\limits_y \mu(y,x)$. Then reiterating Lemma 4 from~\cite{slivkins2014contextual}, we have the following upper bound on the suboptimality of bid $\arm{t}$ chosen at time $t$.
\begin{equation}\label{eq:subopt}
    \Delta(\arm{t},\context{t})\leq 14r(B^\text{sel}_t),
\end{equation}
where $B^\text{sel}_t$ is the ball selected at time $t$ for sampling the bid. Now, if the selected ball satisfied the activation rule, then we have a similar but enhanced upper bound on the expected reward, mentioned as Corollary 5 in~\cite{slivkins2014contextual},
\begin{equation}\label{eq:suboptsaturat}
    \Delta(\arm{t},\context{t})\leq 10r(B^\text{sel}_t).
\end{equation}

Now, consider $\simspace_{\mu,r}\subset\simspace$ which contains points with near optimal expected reward defined as
\begin{equation}
    \simspace_{\mu,r} \overset{\Delta}{=} \{(\arm{},\context{})\in\simspace: \Delta(\arm{},\context{}) \leq 10r\},
    \label{eq:nearoptspacedef}
\end{equation}
and denote its $r$-packing number as $N_r(\simspace_{\mu,r})$, referred to as $N_r$ hereafter.

With the above ingredients, we are now ready to construct the regret bound. For a given radius $r=2^{-i},i\in\mathbb{N}$, let $\mathcal{F}_r$ be the collection of all balls of radius $r$ that have been activated throughout the execution of the algorithm. We can partition all the predictions among the activated balls as follows: for each ball $B\in\mathcal{F}_r$, let $S_B$ be a set of rounds corresponding to ball $B$. $S_B$ consists of the round when $B$ was activated and all rounds when $B$ was selected but no new ball was activated. It can be seen that $\lvert S_B \rvert \leq \mathcal{O}(\tfrac{1}{r^2}\log T + W)$, where the first term comes from the definition of confidence radius~\eqref{eq:craddef} and the second term comes from the fact that a ball could have been selected during predict loop while it became saturated during the update loop. Furthermore, since the point may no longer reside within the domain of the ball, no new ball is activated. Consequently, in the worst-case scenario, there can be a maximum of $W$ such points.\par

If ball $B\in\mathcal{F}_r$ is activated, then by~\eqref{eq:suboptsaturat}, its center lies in $\simspace_{\mu,r}$ defined in~\eqref{eq:nearoptspacedef}. By the separation invariant proved in~\cite{slivkins2014contextual}, the centers of the balls in $\mathcal{F}_r$ are at least $r$ distance away from each other. It follows that $\lvert \mathcal{F}_r\rvert \leq N_r(\simspace_{\mu,r})$. Fixing some $r_0\in(0,1)$, note that in each round $t$ when a ball of radius $<r_0$ was selected, regret is $\Delta(y_t,x_t)\leq\mathcal{O}(r_0)$ as shown in~\eqref{eq:subopt}. Hence, the total regret from all such rounds is at most $\mathcal{O}(r_0T)$. Therefore, it can be written as follows:
\begin{align*}
    R(T)&=\sum_{t=1}^T\Delta(y_t,x_t)\\
    &=\mathcal{O}(r_0T)+\sum_{r=2^{-i}: r_0\leq r\leq 1}\sum_{B\in\mathcal{F}_r}\sum_{t\in S_B}\Delta(y_t,x_t)\\
    &\leq\mathcal{O}(r_0T) + \sum_{r=2^{-i}: r_0\leq r\leq 1}\sum_{B\in\mathcal{F}_r}\lvert S_B\rvert \mathcal{O}(r)\\
    &\leq\mathcal{O}\left(r_0T+\sum_{r=2^{-i}: r_0\leq r\leq 1}N_r\left(\tfrac{1}{r}\log T+Wr\right)\right).
\end{align*}
Finally, taking infimum over $r_0$, we get
\begin{align*}
    R(T)& \leq \mathcal{O}\Biggl(\inf\limits_{r_0\in(0,1)}\Biggl(r_0T\nonumber\\
    &\qquad+\sum_{r=2^{-i}: r_0\leq r\leq 1}N_r\left(\tfrac{1}{r}\log T+Wr\right)\Biggr)\Biggr).
\end{align*}

Let us call this regret bound to be an $N_r$-type guarantee, whereas a corresponding dimension-type guarantee exists. We define $r$-packing dimension $d_c$ corresponding to the $r$-packing number $N_r$ as
\begin{equation*}
    d_c \overset{\Delta}{=} \inf\{d>0: N_r\leq cr^{-d}\quad\forall \in (0,1)\}.
\end{equation*}
Using $i_0=\lceil\tfrac{\log T}{(d_c+2)\log 2}\rceil$ corresponding to $r_0=T^{-1/(d_c+2)}$,
\begin{align*}
    &R(T)\leq\mathcal{O}\left( T^{\tfrac{-1}{d_c+2}}T + \sum_{i=0}^{i_0}c2^{id_c}\left(2^i \log T + W2^{-i}\right) \right)\\
    &=\mathcal{O}\left( T^{\tfrac{d_c+1}{d_c+2}} + c\log T\sum_{i=0}^{i_0-1}2^{i(d_c+1)} + cW\sum_{i=0}^{i_0-1} 2^{i(d_c-1)} \right)\\
    &=\mathcal{O}\left( T^{\tfrac{d_c+1}{d_c+2}} + cT^{\tfrac{d_c+1}{d_c+2}}\log T + cWT^{\tfrac{d_c-1}{d_c+2}}\right)\\
    &=\mathcal{O}\left(T^{\tfrac{d_c+1}{d_c+2}}\log T + WT^{\tfrac{d_c-1}{d_c+2}}\right)\addtocounter{equation}{1}\tag{\theequation}.
\end{align*}
}
This concludes the proof. $\hfill \square$
\vfill

\end{document}